\PassOptionsToPackage{table}{xcolor}
\documentclass[10pt,conference]{IEEEtran}
\IEEEoverridecommandlockouts
\usepackage{cite}
\usepackage{amsmath,amssymb,amsfonts}
\usepackage{algorithmic}
\usepackage{graphicx}
\usepackage{textcomp}

\usepackage{adjustbox}
\usepackage{booktabs}
\usepackage{multirow}
\usepackage{xcolor} 
\usepackage{subcaption}
\usepackage[
  colorlinks=true,
  linkcolor=black, 
  citecolor=black, 
  urlcolor=blue    
]{hyperref}
\usepackage{tikz}

\newcommand\copyrighttext{%
  \footnotesize \textcopyright 2026 IEEE. Personal use of this material is permitted.
  Permission from IEEE must be obtained for all other uses, in any current or future
  media, including reprinting/republishing this material for advertising or promotional
  purposes, creating new collective works, for resale or redistribution to servers or
  lists, or reuse of any copyrighted component of this work in other works.}
\newcommand\copyrightnotice{%
\begin{tikzpicture}[remember picture,overlay]
\node[anchor=south,yshift=10pt] at (current page.south) 
  {\fbox{\parbox{\dimexpr\textwidth-\fboxsep-\fboxrule\relax}{\copyrighttext}}};
\end{tikzpicture}%
}

\def\BibTeX{{\rm B\kern-.05em{\sc i\kern-.025em b}\kern-.08em
    T\kern-.1667em\lower.7ex\hbox{E}\kern-.125emX}}
\begin{document}

\title{GP-Adapter: Gaussian Process CLIP-Adapter for Few-Shot Out-of-Distribution Detection}

\author{\IEEEauthorblockN{1\textsuperscript{st} Taisei Saito}
\IEEEauthorblockA{\textit{Ricoh Company, Ltd.} \\
Ebina, Japan \\
Taisei.Saito1@jp.ricoh.com}
\and
\IEEEauthorblockN{2\textsuperscript{nd} Koretaka Ogata}
\IEEEauthorblockA{\textit{Ricoh Company, Ltd.} \\
Ebina, Japan \\
koretaka.ogata@jp.ricoh.com}
\and
\IEEEauthorblockN{3\textsuperscript{rd} Takafumi Hiroi}
\IEEEauthorblockA{\textit{Ricoh Company, Ltd.} \\
Ebina, Japan \\
takafumi.hiroi@jp.ricoh.com}
}

\maketitle
\copyrightnotice
\begin{abstract}
We propose GP-Adapter, a training-free framework that augments CLIP (Contrastive Language-Image Pre-training) with Gaussian Process (GP) uncertainty modeling for few-shot classification and out-of-distribution (OOD) detection.
While CLIP achieves strong zero-shot recognition, it yields deterministic similarity scores and offers limited uncertainty information, 
which is critical under distribution shift and data scarcity. 
GP-Adapter constructs modality-specific, class-wise one-class GPs on top of frozen CLIP embeddings using an RBF kernel for image features and a linear kernel for text prompts 
and fuses their predictive statistics to produce a variance-aware confidence score for OOD detection. 
The method requires no fine-tuning of the CLIP backbone and relies only on a small $K$-shot cache and lightweight hyperparameter selection, with memory cost scaling as $O(CK^2)$ for $C$ classes and $K$ shots.
Experiments on ImageNet and multiple OOD benchmarks show that GP-Adapter provides competitive few-shot performance and consistently improves OOD detection when combined with prompt-learning baselines, 
highlighting the complementarity between GP-based uncertainty modeling and prompt learning. 
Overall, our results suggest that integrating probabilistic inference with large pre-trained vision-language models can improve reliability in low-data and distribution-shifted settings.
Code is available at \url{https://github.com/tms-byte/GP-Adapter}
\end{abstract}


\section{Introduction}
\label{sec:intro}

Modern deep learning systems are increasingly deployed in open-world environments,
where distribution shifts are inevitable and reliable operation requires the ability to detect out-of-distribution (OOD) inputs that differ from the training data.
In many practical applications such as medical imaging, industrial inspection, and autonomous systems,
models frequently encounter OOD samples, and failing to detect such samples may lead to unreliable or unsafe behavior.
These considerations make OOD detection a fundamental requirement for trustworthy AI systems~\cite{lu2025oodSurvey}.
In addition, labeled training data is often severely limited in practice.
This combination of limited data and distribution shift poses a significant challenge:
with few labeled examples, decision boundaries become highly uncertain,
and models are prone to overconfident predictions on both in-distribution (ID) and OOD inputs.
As a result, principled uncertainty estimation is essential for robust OOD detection in such settings\cite{gal2016dropout}.

Vision-language models (VLMs) such as CLIP (Contrastive Language--Image Pre-training)~\cite{radford2021clip}
have demonstrated strong generalization capabilities by aligning images and text in a shared embedding space learned from large-scale web data.
However, CLIP relies on deterministic similarity scores and lacks explicit mechanisms for uncertainty modeling.
Recent CLIP-based approaches for OOD detection and lightweight adaptation
often depend on prompt tuning or gradient-based optimization~\cite{zhou2022coop,miyai2023locoop,wang2023clipn}.
These methods introduce additional learnable parameters,
which can be unstable or prone to overfitting under extremely limited data,
and are not always suitable for deployment scenarios where gradient-based fine-tuning is costly or undesirable.

Bayesian models, such as Gaussian Processes (GPs)~\cite{rasmussen2006gaussian}, offer a principled framework for uncertainty estimation,
particularly well-suited to low-data regimes.
GPs provide closed-form predictive distributions that explicitly capture predictive uncertainty,
making them attractive for OOD detection~\cite{chen2024uncertainty}.
However, integrating GP-based inference with large-scale multimodal models like CLIP
in an efficient and training-free manner remains an open challenge.

To address these issues, we propose \textbf{GP-Adapter}, a training-free framework that augments frozen CLIP representations with GP-based uncertainty modeling
for few-shot classification and OOD detection.
GP-Adapter constructs modality-specific one-class GPs on CLIP-derived image and text embeddings
using a small K-shot labeled cache, and combines their predictive statistics to compute a variance-aware confidence score for OOD detection.
Importantly, all GP parameters are computed analytically or optimized via lightweight grid search,
enabling rapid adaptation in real-world environments where gradient-based fine-tuning is undesirable.
Because each class-specific one-class GP is built from only $K$ support samples for each of the $C$ classes, 
the memory footprint scales as $O(CK^2)$, 
making GP-Adapter scalable to large-scale datasets such as ImageNet-1k~\cite{deng2009imagenet}.
Moreover, our results show that GP-based uncertainty modeling is complementary to prompt-learning approaches: 
combining GP-Adapter with prompt-tuning baselines consistently strengthens OOD detection while preserving ID accuracy.

The contributions of our work are summarized as follows:
\begin{itemize}
  \item We propose GP-Adapter, a training-free framework that augments frozen CLIP with GP-based uncertainty modeling for OOD detection under severe data scarcity.
  \item We introduce a modality-specific, class-wise one-class GP formulation that models image and text embeddings independently and combines their predictive statistics to produce a variance-aware confidence score.
  \item Extensive experiments demonstrate that GP-Adapter improves few-shot OOD detection performance across multiple benchmarks without requiring additional fine-tuning or OOD datasets, and further gains are obtained when integrating it with prompt-learning methods.
\end{itemize}

\section{Related Work}
\label{sec:related}

Early OOD detection methods are primarily based on confidence scores derived from
the output of trained discriminative neural networks.
Hendrycks and Gimpel~\cite{hendrycks2017baseline} proposed the Maximum Softmax Probability (MSP),
while Liu \emph{et al.}~\cite{liu2020energy} introduced energy-based scores to improve
OOD detection performance.
Other approaches exploit feature-space information.
ViM~\cite{wang2022vim} models the in-distribution feature subspace, and deep nearest
neighbor methods~\cite{sun2022deepnn} detect OOD samples based on distances in the
embedding space.
Uncertainty-based approaches leverage Bayesian neural networks (BNNs)\cite{gal2016dropout} or Gaussian Processes (GPs)\cite{rasmussen2006gaussian} to estimate predictive uncertainty for OOD detection.
Prior work has shown that Bayesian inference enables neural networks to
assign higher uncertainty to out-of-distribution samples, 
providing a useful signal for OOD detection\cite{e107-d_8_949,10650576,chen2024uncertainty}.

Recent works have explored OOD detection in the context of CLIP without retraining the backbone.
MCM~\cite{ming2022mcm} utilizes the maximum class probability as an OOD score,
demonstrating that CLIP's zero-shot predictions already contain useful signals for OOD detection.
GL-MCM~\cite{miyai2025glmcm} further extends this idea by incorporating global feature statistics into the MCM score.
Tip-Adapter~\cite{zhang2022tipadapter} proposes a training-free, cache-based few-shot adaptation
mechanism for improving in-distribution (ID) classification, 
and an extension for OOD detection has also been proposed~\cite{chen2024dualadapter}. 
Prompt-learning approaches have also been adapted for OOD detection.
CoOp~\cite{zhou2022coop} introduces learnable prompt vectors to adapt CLIP in few-shot settings via gradient-based optimization.
Building upon CoOp, LoCoOp~\cite{miyai2023locoop} incorporates background-aware regularization to explicitly improve OOD detection performance.
Other methods enhance OOD detection through alternative mechanisms.
CLIPN~\cite{wang2023clipn} introduces explicit negative textual concepts for zero-shot
OOD detection by adding a negative text encoder trained with ``No'' prompts, 
NPOS~\cite{tao2023npos} leverages synthetic outliers generated in the feature space during training, and
SeTAR~\cite{li2024setar} employs selective low-rank approximation to suppress
non-discriminative components in CLIP representations.
In contrast, our work introduces a Gaussian Process-based uncertainty modeling layer on top of CLIP embeddings, 
enabling principled OOD detection and few-shot classification without gradient-based training or additional OOD data.

\begin{figure}[t]
  \centering
  \includegraphics[width=0.72\columnwidth]{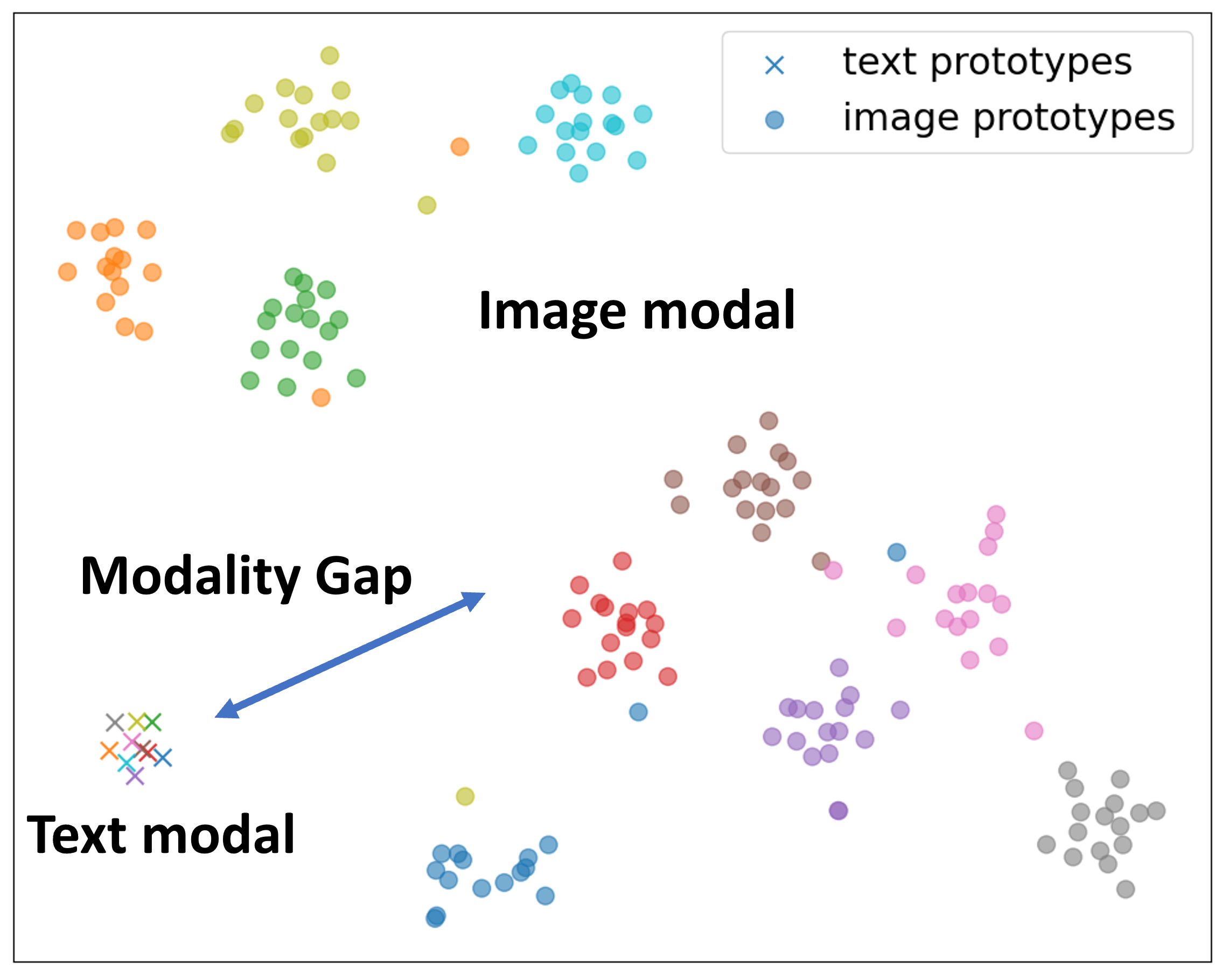}
  \caption{
  t\mbox{-}SNE visualization of CLIP embeddings.
  Image and text embeddings exhibit distinct clustering patterns
  with a systematic offset, suggesting differences in their geometric structure
  within the shared embedding space.
  }
  \label{fig:modality-gap}
  \vspace{-0.5em}
\end{figure}

\begin{figure*}[ht]
  \centering
  \includegraphics[width=0.95\textwidth]{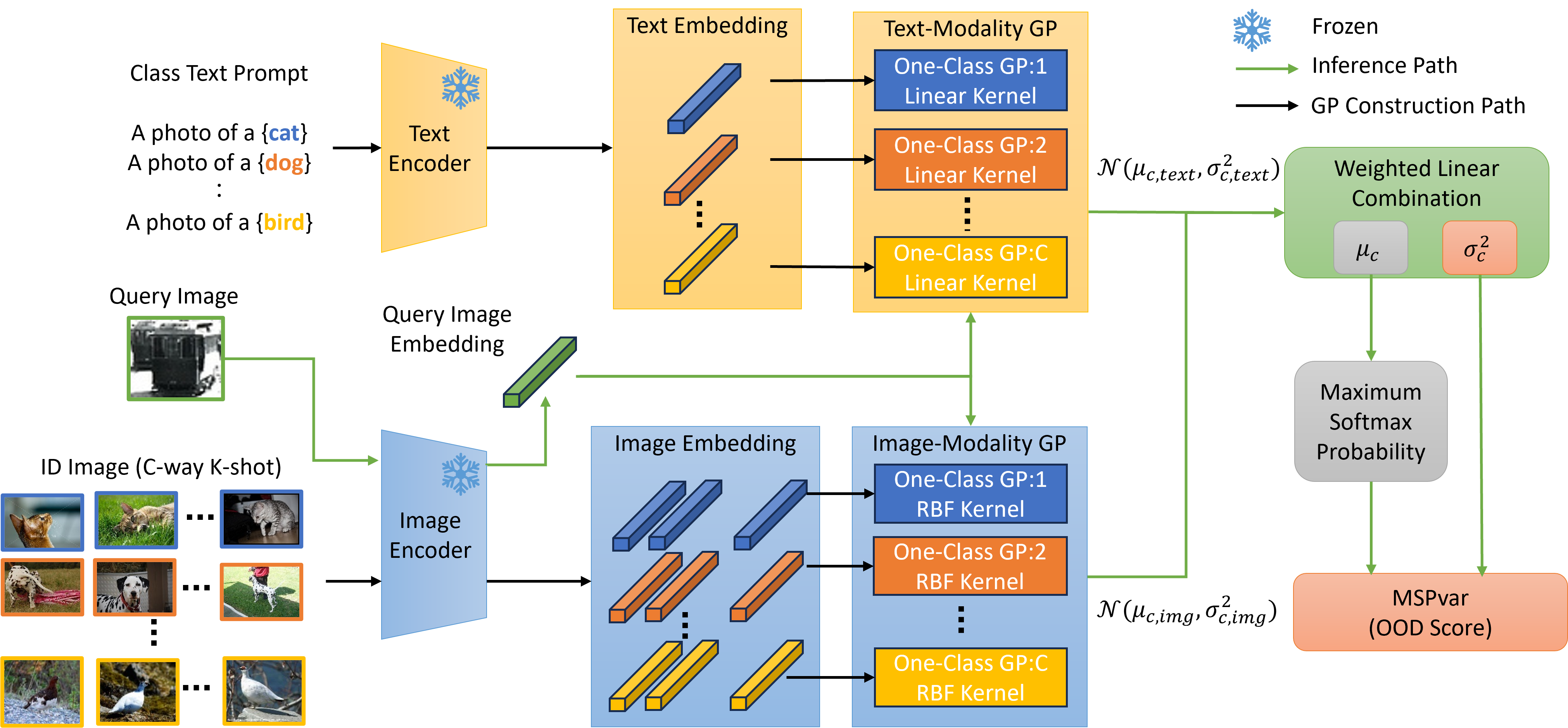}
  \caption{\textbf{{Overview of GP-Adapter with one-class Gaussian Processes.}}
  Frozen CLIP encoders produce image embeddings from a small $K$-shot support set and a query image, and text embeddings from class-specific text prompts.
  For each class, modality-specific one-class Gaussian Processes output predictive means and variances.
  These are fused via a weighted combination and converted into OOD scores using the proposed variance-aware MSP.}
  \label{fig:overview}
  \vspace{-0.6em}
\end{figure*}

\section{Preliminaries}
\subsection{Gaussian Process Regression}
A Gaussian Process is a non-parametric Bayesian model that defines a distribution over functions.
It is characterized by a mean function \( m(\mathbf{x}) \) and a covariance function \( k(\mathbf{x}, \mathbf{x}') \), 
also known as kernel, which defines the similarity between input points. 
The kernel function plays a central role in controlling the smoothness and generalization behavior of the function by encoding prior assumptions about the function's structure.
Formally, a GP is written as:
\[
f(\mathbf{x}) \sim \mathcal{GP}(m(\mathbf{x}), k(\mathbf{x}, \mathbf{x}')).
\]

Given training data \( \mathcal{D} = \{(\mathbf{x}_i, y_i)\}_{i=1}^n \), where \( \mathbf{x}_i \in \mathbb{R}^d \) are input features and \( y_i \in \mathbb{R} \) are noisy observations of the latent function \( f(\mathbf{x}_i) \), we assume the following observation model:
\[
y_i = f(\mathbf{x}_i) + \epsilon_i, \quad \epsilon_i \sim \mathcal{N}(0, \sigma^2)
\]
where \( \sigma^2 \) is the variance of the independent Gaussian noise added to each observation.

Under this model, the joint distribution of the training outputs and the function value at a test point \( \mathbf{x}_* \) is given by:
\[
\begin{bmatrix}
\mathbf{y} \\
f_*
\end{bmatrix}
\sim \mathcal{N} \left(
\mathbf{0},
\begin{bmatrix}
K + \sigma^2 I & k_* \\
k_*^\top & k_{**}
\end{bmatrix}
\right).
\]
Here, \( K \in \mathbb{R}^{n \times n} \) denotes the kernel matrix computed over the training inputs, 
where each entry \( K_{ij} = k(\mathbf{x}_i, \mathbf{x}_j) \). 
The vector \( k_* \in \mathbb{R}^n \) represents the kernel evaluations between the training inputs and the test input \( \mathbf{x}_* \), 
such that \( (k_*)_i = k(\mathbf{x}_i, \mathbf{x}_*) \). 
The scalar \( k_{**} = k(\mathbf{x}_*, \mathbf{x}_*) \) is the kernel value at the test input itself.

The predictive distribution at test point \( \mathbf{x}_* \) is Gaussian with mean and variance:
\begin{align}
\mu_* &= k_*^\top (K + \sigma^2 I)^{-1} \mathbf{y}, \label{eq:mu}\\
\sigma^2_* &= k_{**} - k_*^\top (K + \sigma^2 I)^{-1} k_*. \label{eq:sigma}
\end{align}

This formulation allows GPs to provide not only point predictions, computed as the predictive mean, 
but also uncertainty estimates, computed by the predictive variance, 
which are crucial for tasks such as few-shot learning and out-of-distribution (OOD) detection. 
A larger predictive variance indicates lower confidence in the prediction,
whereas a smaller variance indicates higher confidence.

\subsection{CLIP Representations}
CLIP is a multimodal representation learning framework that embeds images and text
into a shared semantic space using separate modality-specific encoders.

Given an image \( \mathbf{x}_{\text{img}} \) and a set of candidate class descriptions \( \{ \mathbf{t}_1, \dots, \mathbf{t}_C \} \), 
CLIP produces $\ell_2$-normalized feature vectors \( \mathbf{f}_{\text{img}} \in \mathbb{R}^d \) and \( \mathbf{f}_{\text{text}, c} \in \mathbb{R}^d \) for each class \( c \). 
The prediction score for class \( c \) is computed using cosine similarity:
\begin{equation}
s_c = \mathbf{f}_{\text{img}}^\top \mathbf{f}_{\text{text}, c}.
\end{equation}
The predicted class is then given by:
\begin{equation}
\hat{c} = \arg\max_{c} \, s_c.
\end{equation}
These scores can be interpreted as similarity-based logits and are commonly used for zero-shot classification.

Although CLIP aligns image and text embeddings in a shared space, 
Fig.~\ref{fig:modality-gap} suggests a residual cross-modal misalignment, 
where the image and text embeddings form distinct clusters and exhibit a systematic offset.
This observation highlights that, despite semantic alignment,
the two modalities exhibit different geometric characteristics in the embedding space.

\section{METHODOLOGY}
\subsection{Problem Setup}\label{sec:setup}
We consider a few-shot classification setting with $C$ classes and a small labeled cache of $K$ examples per class, denoted as
$\mathcal{S} = \{(x_i, y_i)\}_{i=1}^{CK}$, where $x_i$ is an image and $y_i \in \{1,\dots,C\}$ is its class label.
Given a query image $x_q$, our objective is to predict its class label for ID classification and leverage predictive uncertainty as a signal for OOD detection.

\subsection{Overview of our GP-Adapter}\label{sec:overview}
Fig.~\ref{fig:overview} illustrates the architecture of \textbf{GP-Adapter}, which augments CLIP with GP-based Bayesian inference for few-shot OOD detection.
The pipeline combines CLIP's multimodal representations with GP-based uncertainty estimation to enable uncertainty-aware predictions under data scarcity and distribution shift.

We begin by constructing a K-shot support set for each class $c$.
For the image modality, we encode these C-way K-shot labeled images using CLIP's image encoder to obtain image embeddings.
For the text modality, we use class-specific text prompts (e.g., “a photo of a \{class\}”) and encode them using CLIP's text encoder to obtain text embeddings.
Following~\cite{radford2021clip}, we adopt a prompt-ensemble strategy and use the standard set of seven ImageNet templates.
These embeddings are then used to construct modality-specific kernel matrices:
an RBF kernel for image embeddings and a linear kernel for text embeddings,
which are used to construct independent one-class GPs for each modality.

At inference time, the frozen CLIP image encoder encodes the query image $x_q$ into an $\ell_2$-normalized embedding $z_q = f_{\text{img}}(x_q)$.
For each class $c$, we construct two independent one-class GP regressors, 
one for the image modality and one for the text modality.
Given the original multi-class labels $y_i \in \{1,\dots,C\}$, 
we assign a positive regression target $\tilde{y}_i^{(c)} = 1$ to all support samples belonging to class $c$ when constructing the one-class GPs for each modality.

Using these targets, the two GPs produce predictive distributions over a latent class score $s_c(z_q)$ for the query embedding:
\begin{align*}
p\!\left(s_c(z_q)\mid z_q\right)_{\text{img}} &= \mathcal{N}\!\left(\mu_{c,\text{img}}(z_q), \sigma^2_{c,\text{img}}(z_q)\right),\\
p\!\left(s_c(z_q)\mid z_q\right)_{\text{text}} &= \mathcal{N}\!\left(\mu_{c,\text{text}}(z_q), \sigma^2_{c,\text{text}}(z_q)\right).
\end{align*}
The predictive mean and variance are computed using the standard GP regression formulas given in \eqref{eq:mu} and \eqref{eq:sigma}.
For the image modality, the GP prediction is obtained by computing kernel values between the query image embedding and the support image embeddings.
For the text modality, the query image embedding is compared to the fixed text embeddings via the linear kernel,
leveraging the alignment property of CLIP's embedding space, and the GP prediction is obtained using the
corresponding kernel values and the precomputed Gram matrix among text embeddings.

Assuming independence between the two modalities, 
we fuse their predictive distributions via a weighted linear combination.
The resulting predictive mean and variance for class $c$ are given by:
\begin{align*}
\mu_c(z_q) &= \alpha \mu_{c,\text{img}}(z_q) + (1-\alpha)\mu_{c,\text{text}}(z_q), \\
\sigma_c^2(z_q) &= \alpha^2\sigma^2_{c,\text{img}}(z_q) + (1-\alpha)^2\sigma^2_{c,\text{text}}(z_q),
\end{align*}
where $\alpha \in [0,1]$ controls the contribution of each modality.
In our experiments, we set \( \alpha = 0.15 \) unless otherwise specified.

We convert the fused predictive means $\{\mu_c(z_q)\}_{c=1}^C$ into class probabilities using the softmax function:
\begin{equation}
p(y=c \mid z_q) = \frac{\exp(\mu_c(z_q))}
{\sum_{c'=1}^C \exp(\mu_{c'}(z_q))}.
\end{equation}

Following the Maximum Softmax Probability (MSP) criterion,
the confidence score for a query sample is defined as:
\begin{equation}
\mathrm{MSP}(z_q) = \max_{c} \, p(y=c \mid z_q).
\end{equation}

\begin{figure}[t]
  \centering
  \begin{minipage}{0.48\linewidth}
    \centering
    \includegraphics[width=\linewidth]{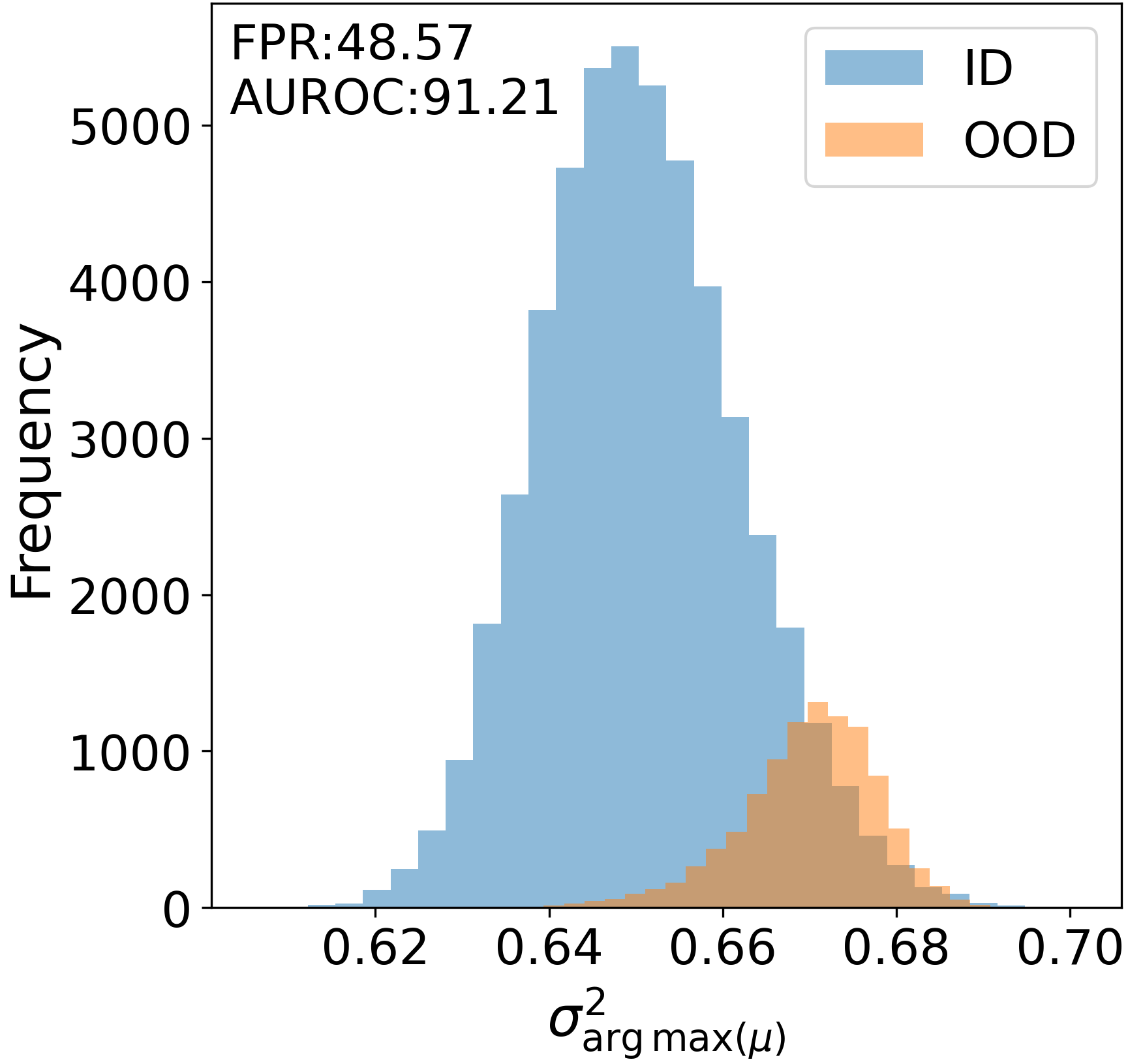}
    {\small (a) Predicted-class variance}
  \end{minipage}\hfill
  \begin{minipage}{0.48\linewidth}
    \centering
    \includegraphics[width=\linewidth]{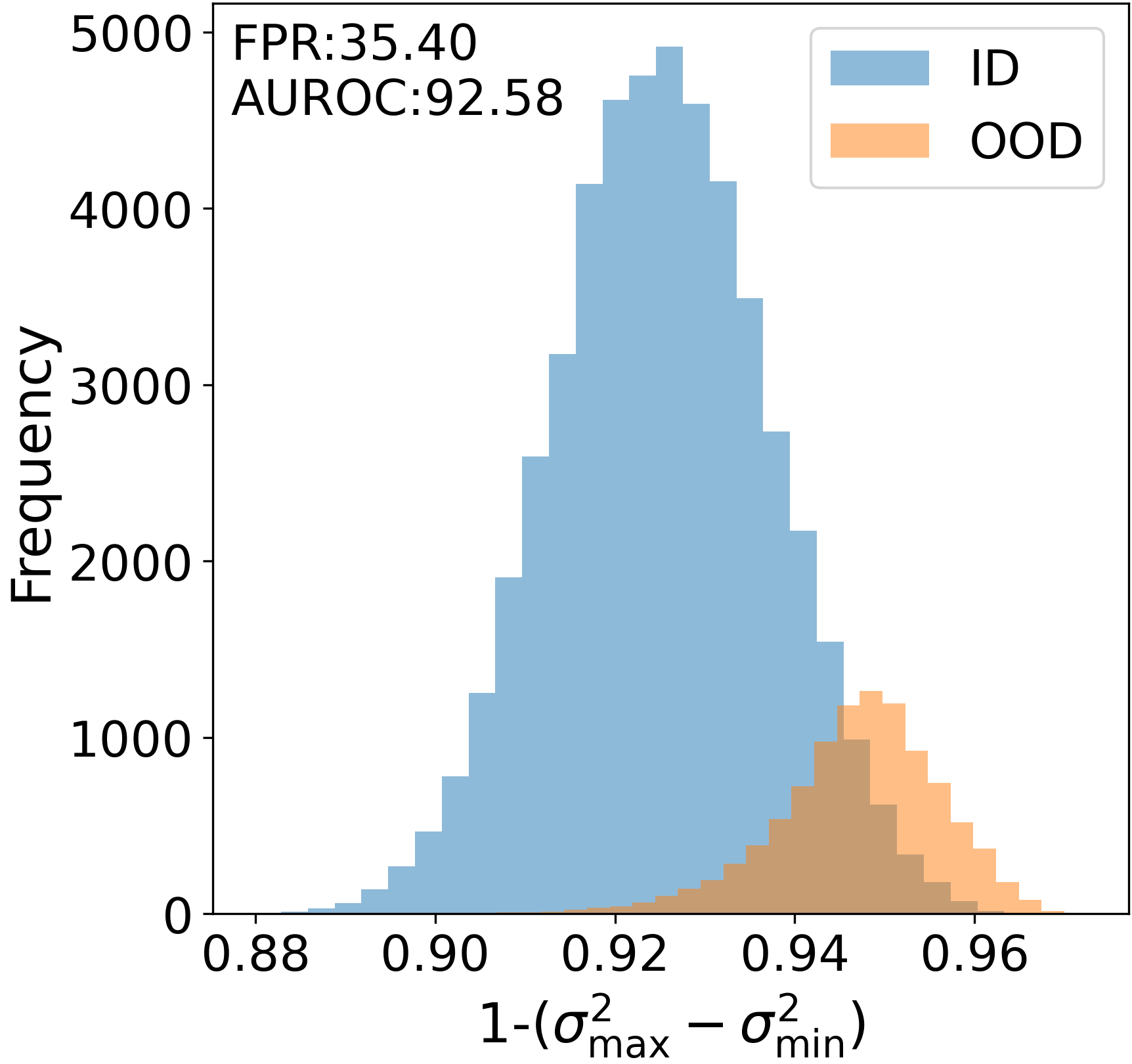}
    {\small (b) Max-min variance range}
  \end{minipage}
  \caption{Range of GP predictive variances across classes improves separation between ID and OOD samples.}
  \label{fig:uncertainty-hist}
\end{figure}

To incorporate predictive uncertainty, we further define a normalized variance-aware confidence score that accounts for the range of predictive variances across classes:
\begin{equation}
\mathrm{MSP}_{\text{var}}(z_q)
= \mathrm{MSP}(z_q)
\left(
1 + \frac{\sigma^2_{\max}(z_q) - \sigma^2_{\min}(z_q)}
{\sigma^2_{\max}(z_q) + \sigma^2_{\min}(z_q)}
\right),
\label{eq:mspvar}
\end{equation}
where $\sigma^2_{\max}(z_q)$ and $\sigma^2_{\min}(z_q)$ represent the maximum and minimum predictive variances across classes, respectively.
Lower $\text{MSP}_{\text{var}}$ values indicate higher uncertainty and are used to identify OOD inputs.
For OOD detection, we apply a threshold $\lambda$: if $\mathrm{MSP}_{\text{var}}(z_q) \ge \lambda$, the sample is detected as ID; otherwise, detected as OOD.
This design is motivated by empirical observations that the variance range across classes provides a stronger signal for separating ID and OOD samples than the variance of the most confident class alone.
Since GP predictive variance is typically lower near the support set and higher far from it, ID queries yield a wide max--min variance range, 
whereas OOD queries tend to produce more uniformly high variances and thus a narrower range.
As illustrated in Fig.~\ref{fig:uncertainty-hist}, incorporating class-wise variance range leads to improved separation between ID and OOD distributions.

\subsection{Kernel Selection}\label{sec:kernel}
A key design choice in GP-Adapter is the use of modality-specific kernels, 
which is motivated by the geometric properties of CLIP embeddings.
Although CLIP aligns image and text representations in a shared embedding space,
the two modalities exhibit different statistical characteristics and residual misalignment, 
as illustrated in Fig.~\ref{fig:modality-gap}.
This observation suggests that applying a single distance-sensitive kernel uniformly across modalities may lead to distorted similarity estimates.

For the image modality, we use the Radial Basis Function (RBF) kernel, which assumes local smoothness and captures non-linear relationships in the feature space. 
In CLIP, image embeddings exhibit a geometry where Euclidean distance
provides a meaningful measure of visual similarity, making the RBF kernel a natural choice.
The kernel is defined as:
\begin{equation}
k_{\text{img}}(\mathbf{z}, \mathbf{z}') = \exp\left(-\frac{\|\mathbf{z} - \mathbf{z}'\|^2}{2\theta^2}\right),
\end{equation}
where \( \theta \) is the length-scale parameter.

For the text modality, we apply a linear kernel on $\ell_2$-normalized prompt embeddings.
Since CLIP computes image-text similarity using cosine similarity, the inner product between $\ell_2$-normalized embeddings is directly aligned with CLIP's standard scoring mechanism.
This choice avoids introducing an additional distance metric that may amplify residual cross-modal misalignment.
Formally, the text kernel is given by:
\begin{equation}
k_{\text{text}}(\mathbf{t}, \mathbf{t}')
= \mathbf{t}^\top \mathbf{t}',
\end{equation}
and the cross-covariance between a query image embedding $\mathbf{z}_q$
and a text embedding $\mathbf{t}$ is computed as
$k_{\text{text}}(\mathbf{t}, \mathbf{z}_q) = \mathbf{t}^\top \mathbf{z}_q$.

By restricting distance-sensitive kernels to the image modality and using an inner-product-based kernel for text prompts,
GP-Adapter respects the modality-specific geometry of CLIP embeddings.
This design reduces sensitivity to cross-modal misalignment and 
leads to more stable similarity estimation and uncertainty modeling in OOD scenarios.

\subsection{Learning Kernel Hyperparameters}\label{sec:hyper}
To effectively model uncertainty in few-shot settings while maintaining scalability,
we optimize kernel hyperparameters independently for each one-class Gaussian Process.
Specifically, for each class $c$, a separate GP is constructed using only
its $K$-shot support samples.

Accordingly, the training inputs for class $c$ are given by
$\mathbf{X}^{(c)} \in \mathbb{R}^{K \times d}$, consisting of $K$ support embeddings,
and the corresponding positive regression targets
$\tilde{\mathbf{y}}^{(c)} \in \mathbb{R}^{K}$ are defined as in Sec.~\ref{sec:overview}.
For each class-specific GP, kernel hyperparameters are selected by maximizing
the log marginal likelihood:
\begin{align*}
\log p(\tilde{\mathbf{y}}^{(c)} \mid \mathbf{X}^{(c)})
= &-\frac{1}{2} \tilde{\mathbf{y}}^{(c)\top}
\bigl(K^{(c)} + \sigma^2 I\bigr)^{-1} \tilde{\mathbf{y}}^{(c)}\\
&- \frac{1}{2} \log \bigl|K^{(c)} + \sigma^2 I\bigr|
- \frac{K}{2} \log 2\pi ,
\end{align*}
where $K^{(c)}$ denotes the kernel matrix computed from $\mathbf{X}^{(c)}$.

In practice, we assume a nearly noise-free observation model and fix the noise
variance to $\sigma^2 = 10^{-6}$ for numerical stability.
We parameterize the kernel matrix as:
\begin{equation}
K^{(c)}(\mathbf{X}^{(c)}, \mathbf{X}^{(c)})
= \sigma_f^{2(c)} K_0(\mathbf{X}^{(c)}, \mathbf{X}^{(c)} \mid \theta^{(c)}),
\end{equation}
where $\sigma_f^{2(c)}$ is the signal variance and $\theta^{(c)}$ denotes
the kernel length-scale.
For a fixed $\theta^{(c)}$, the optimal signal variance admits a closed-form solution:
\begin{equation}
\hat{\sigma}_f^{2(c)}
= \frac{1}{K} \tilde{\mathbf{y}}^{(c)\top}
K_0(\mathbf{X}^{(c)}, \mathbf{X}^{(c)} \mid \theta^{(c)})^{-1}
\tilde{\mathbf{y}}^{(c)} .
\label{eq:sigf}
\end{equation}

For the image modality, the length-scale $\theta^{(c)}$ of the RBF kernel is selected via grid search, 
while the signal variance $\sigma_f^{2(c)}$ of the RBF kernel is updated in closed form for each candidate length-scale using \eqref{eq:sigf}, 
resulting in a profile likelihood optimization.
For the text modality, the signal variance $\sigma_f^{2(c)}$ of the linear kernel is updated in closed form without iterative search.

To mitigate overfitting under extremely limited data, we impose an upper bound
$\tau$ on the log marginal likelihood during length-scale selection:
candidate length-scales whose log marginal likelihood exceeds $\tau$ are excluded,
and we select the best length-scale among the remaining candidates.
That is, we maximize the log marginal likelihood subject to
$\log p(\tilde{\mathbf{y}}^{(c)} \mid \mathbf{X}^{(c)}) \le \tau$.
Unless otherwise specified, we set $\tau=-5$ in all experiments.

Importantly, since each class-specific GP is built from only $K$ support samples,
the memory footprint for class-wise kernel matrices scales as $O(CK^2)$ rather than $O(N^2)$,
where $C$ denotes the number of classes and $N$ the total number of training samples.
This design enables scalable GP-based uncertainty modeling even for large-scale
datasets such as ImageNet-1k under few-shot settings.

\begin{table*}[tb]
  \centering
  \caption{Comparison of OOD detection performance with ImageNet-1k as ID data. $\downarrow$ indicates lower is better, $\uparrow$ indicates higher is better. $^{\dagger}$ indicates results reported in \cite{tao2023npos}, $^{\ddagger}$ indicates results reported in \cite{wang2023clipn}, and $^{\dagger\ddagger}$ indicates results reported in \cite{li2024setar}.}
  \label{tab:imagenet-1k-ood}
  \begin{adjustbox}{max width=\textwidth}
  \begin{tabular}{lcccccccccccc}
  \toprule
  \multirow{2}{*}{\textbf{Method}} & 
  \multicolumn{2}{c}{\textbf{iNaturalist}} & 
  \multicolumn{2}{c}{\textbf{SUN}} & 
  \multicolumn{2}{c}{\textbf{Places}} & 
  \multicolumn{2}{c}{\textbf{Texture}} & 
  \multicolumn{2}{c}{\textbf{Average}} \\
  \cmidrule(lr){2-3} \cmidrule(lr){4-5} \cmidrule(lr){6-7} \cmidrule(lr){8-9} \cmidrule(lr){10-11}
    & FPR95$\downarrow$ & AUROC$\uparrow$ & FPR95$\downarrow$ & AUROC$\uparrow$ & 
      FPR95$\downarrow$ & AUROC$\uparrow$ & FPR95$\downarrow$ & AUROC$\uparrow$ & 
      FPR95$\downarrow$ & AUROC$\uparrow$ \\
  \midrule
  
  \multicolumn{11}{l}{\textit{Zero-shot}} \\
  MCM~\cite{ming2022mcm} & 30.94 & 94.61 & 37.67 & 92.56 & 44.76 & 89.76 & 57.91 & 86.10 & 42.82 & 90.76 \\
  GL-MCM~\cite{miyai2025glmcm} & 15.18 & 96.71 & 30.42 & 93.09 & 38.85 & 89.90 & 57.93 & 83.63 & 35.47 & 90.83 \\
  SeTAR~\cite{li2024setar}$^{\dagger\ddagger}$ & 26.92 & 94.67 & 35.57 & 92.79 & 42.64 & 90.16 & 55.83 & 86.58 & 40.24 & 91.05 \\
  \midrule
  
  \multicolumn{11}{l}{\textit{Fine-tuned}} \\
  MSP~\cite{fort2021exploring}$^{\dagger}$ & 54.05 & 87.43 & 73.37 & 78.03 & 72.98 & 78.03 & 68.85 & 79.06 & 67.31 & 80.64\\ 
  ODIN~\cite{liang2018odin}$^{\dagger}$ & 30.22 & 94.65 & 54.04 & 87.17 & 55.06 & 85.54 & 51.67 & 87.85 & 47.75 & 88.80 \\
  ViM~\cite{wang2022vim}$^{\dagger}$ & 32.19 & 94.65 & 54.01 & 87.19 & 60.67 & 83.75 & 53.94 & 87.17 & 50.45 & 88.19 \\
  KNN~\cite{sun2022deepnn}$^{\dagger}$ & 29.17 & 91.52 & 52.62 & 86.12 & 57.02 & 84.43 & 54.93 & 87.12 & 48.44 & 87.30 \\
  NPOS~\cite{tao2023npos}$^{\dagger}$ & 16.58 & 96.19 & 43.77 & 90.44 & 49.62 & 87.54 & 43.65 & 89.10 & 38.91 & 90.82 \\
  CLIPN~\cite{wang2023clipn}$^{\ddagger}$ & 23.94 & 95.27 & \textbf{26.17} & \textbf{93.93} & \textbf{33.45} & 90.93 & 40.83 & \textbf{92.28} & 31.10 & 93.10 \\
  \midrule
  
  \multicolumn{11}{l}{\textit{Few-shot (16-shot)}} \\
  CoOp~\cite{zhou2022coop} & 29.63 & 93.59 & 35.33 & 92.42 & 42.31 & 89.89 & 41.58 & 90.59 & 37.21 & 91.62 \\
  LoCoOp\cite{miyai2023locoop} & 23.78 & 95.20 & 30.75 & 93.89 & 37.27 & 91.24 & 41.70 & 91.05 & 33.38 & 92.85 \\
  \rowcolor{gray!15}
  GP-Adapter (ours) & 15.78 & 96.78 & 37.54 & 92.36 & 42.32 & 90.37 & 48.52 & 89.37 & 36.04 & 92.12 \\
  \rowcolor{gray!15}
  GP-Adapter+CoOp (ours) & 16.24 & 96.53 & 32.78 & 93.02 & 37.27 & \textbf{91.58} & \textbf{39.27} & 91.74 & 31.39 & 93.22 \\
  \rowcolor{gray!15}
  GP-Adapter+LoCoOp (ours) & \textbf{12.66} & \textbf{97.23} & 31.52 & 93.60 & 37.28 & 91.56 & 39.47 & 91.94 & \textbf{30.23} & \textbf{93.58} \\
  \bottomrule
  \end{tabular}
  \end{adjustbox}
\end{table*}

\begin{table}[tb]
  \centering
  \caption{Comparison in ID accuracy on ImageNet-1k validation data.}
  \label{tab:id-accuracy}
  \begin{tabular}{lc}
  \toprule
  \textbf{Method} & \textbf{Top-1 Accuracy} \\
  \midrule
  MCM    & 68.53 \\
  CoOp   & 71.93 \\
  LoCoOp & 71.63 \\
  \rowcolor{gray!15}
  GP-Adapter(ours) & 70.25 \\
  \rowcolor{gray!15}
  GP-Adapter+CoOp(ours) & \textbf{72.10} \\
  \rowcolor{gray!15}
  GP-Adapter+LoCoOp(ours) & 71.51 \\
  \bottomrule
  \end{tabular}
\end{table}

\section{Experiments}
\label{sec:exp}

\begin{table}[t]
  \centering
  \caption{Effect of upper bound $\tau$ on OOD detection performance (evaluated under GP-Adapter+LoCoOp).}
  \label{tab:ablation-tau}
  \small
  \begin{tabular}{lcc}
  \toprule
  $\tau$ & FPR95$\downarrow$ & AUROC$\uparrow$ \\
  \midrule
  $\infty$ & 33.12 & 92.93 \\
  5      & 33.29 & 93.02 \\
  0      & \textbf{29.99} & 93.55 \\
  -5     & 30.23 & \textbf{93.58} \\
  -10    & 33.43 & 93.08 \\
  \bottomrule
  \end{tabular}
\end{table}

\begin{table}[t]
  \centering
  \caption{Effect of modality ratio $\alpha$ on OOD detection performance (evaluated under GP-Adapter+LoCoOp).}
  \label{tab:ablation-alpha}
  \small
  \begin{tabular}{lcc}
  \toprule
  $\alpha$ & FPR95$\downarrow$ & AUROC$\uparrow$ \\
  \midrule
  0.05     & 31.94 & 93.16 \\
  0.1      & 30.46 & 93.47 \\
  0.15     & \textbf{30.23} & \textbf{93.58} \\
  0.2      & 32.09 & 93.29 \\
  \bottomrule
  \end{tabular}
\end{table}

\begin{table*}[tb]
  \centering
  \caption{Comparison of OOD detection performance with ResNet-50 backbone. $\downarrow$ indicates lower is better, $\uparrow$ indicates higher is better.}
  \label{tab:rn50-ood}
  \begin{adjustbox}{max width=\textwidth}
  \begin{tabular}{lcccccccccccc}
  \toprule
  \multirow{2}{*}{\textbf{Method}} & 
  \multicolumn{2}{c}{\textbf{iNaturalist}} & 
  \multicolumn{2}{c}{\textbf{SUN}} & 
  \multicolumn{2}{c}{\textbf{Places}} & 
  \multicolumn{2}{c}{\textbf{Texture}} & 
  \multicolumn{2}{c}{\textbf{Average}} \\
  \cmidrule(lr){2-3} \cmidrule(lr){4-5} \cmidrule(lr){6-7} \cmidrule(lr){8-9} \cmidrule(lr){10-11}
    & FPR95$\downarrow$ & AUROC$\uparrow$ & FPR95$\downarrow$ & AUROC$\uparrow$ & 
      FPR95$\downarrow$ & AUROC$\uparrow$ & FPR95$\downarrow$ & AUROC$\uparrow$ & 
      FPR95$\downarrow$ & AUROC$\uparrow$ \\
  \midrule
  MCM & 31.98 & 93.86 & 46.09 & \textbf{90.75} & 60.56 & 85.67 & 60.00 & 85.72 & 49.66 & 89.00 \\
  CoOp & 34.35 & 93.17 & \textbf{44.97} & 90.60 & 55.63 & 86.35 & \textbf{42.82} & \textbf{89.94} & 44.40 & 90.02 \\
  LoCoOp & 39.86 & 92.62 & 51.55 & 90.23 & 58.66 & 86.34 & 49.07 & 89.18 & 49.79 & 89.60 \\
  \rowcolor{gray!15}
  GP-Adapter (ours) & 29.37 & 94.22 & 46.97 & 90.39 & 56.08 & 86.66 & 55.32 & 87.26 & 46.93 & 89.63 \\
  \rowcolor{gray!15}
  GP-Adapter+CoOp (ours) & \textbf{23.07} & \textbf{95.36} & 47.37 & 89.88 & \textbf{54.26} & \textbf{86.92} & 44.39 & 89.87 & \textbf{42.27} & \textbf{90.51} \\
  \rowcolor{gray!15}
  GP-Adapter+LoCoOp (ours) & 24.25 & 95.25 & 51.40 & 89.72 & 55.96 & 86.76 & 48.53 & 89.56 & 45.03 & 90.32 \\
  \bottomrule
  \end{tabular}
  \end{adjustbox}
\end{table*}

\begin{table*}[tb]
  \centering
  \caption{Comparison of OOD detection performance with ImageNet-100 as ID data. $\downarrow$ indicates lower is better, $\uparrow$ indicates higher is better.}
  \label{tab:imagenet-100-ood}
  \begin{adjustbox}{max width=\textwidth}
  \begin{tabular}{lcccccccccccc}
  \toprule
  \multirow{2}{*}{\textbf{Method}} & 
  \multicolumn{2}{c}{\textbf{iNaturalist}} & 
  \multicolumn{2}{c}{\textbf{SUN}} & 
  \multicolumn{2}{c}{\textbf{Places}} & 
  \multicolumn{2}{c}{\textbf{Texture}} & 
  \multicolumn{2}{c}{\textbf{Average}} \\
  \cmidrule(lr){2-3} \cmidrule(lr){4-5} \cmidrule(lr){6-7} \cmidrule(lr){8-9} \cmidrule(lr){10-11}
    & FPR95$\downarrow$ & AUROC$\uparrow$ & FPR95$\downarrow$ & AUROC$\uparrow$ & 
      FPR95$\downarrow$ & AUROC$\uparrow$ & FPR95$\downarrow$ & AUROC$\uparrow$ & 
      FPR95$\downarrow$ & AUROC$\uparrow$ \\
  \midrule
  MCM & 18.13 & 96.77 & 36.45 & 94.54 & 34.52 & 94.36 & 41.22 & 92.25 & 32.58 & 94.48 \\
  CoOp & 19.77 & 96.55 & 19.69 & 96.27 & 23.78 & 95.22 & 15.95 & 96.91 & 19.80 & 96.24 \\
  LoCoOp & 7.37 & 98.16 & 18.43 & \textbf{96.80} & 22.88 & 95.71 & \textbf{15.07} & \textbf{97.20} & 15.94 & \textbf{96.97} \\
  \rowcolor{gray!15}
  GP-Adapter (ours) & \textbf{2.46} & \textbf{99.19} & 19.09 & 96.62 & \textbf{21.60} & \textbf{96.13} & 23.97 & 95.86 & 16.78 & 96.95 \\
  \rowcolor{gray!15}
  GP-Adapter+CoOp (ours) & 12.89 & 97.44 & \textbf{17.83} & 96.51 & 21.61 & 95.57 & 17.37 & 96.71 & 17.42 & 96.56 \\
  \rowcolor{gray!15}
  GP-Adapter+LoCoOp (ours) & 3.91 & 98.89 & 20.50 & 96.31 & 23.41 & 95.46 & 15.53 & 96.98 & \textbf{15.84} & 96.91 \\
  \bottomrule
  \end{tabular}
  \end{adjustbox}
\end{table*}

\subsection{Experimental Setting}
\textbf{Datasets.}
We evaluate our method on ImageNet-1k\cite{deng2009imagenet} as the ID classification benchmark. 
For OOD detection, we use iNaturalist\cite{vanhorn2018inaturalist}, SUN\cite{xiao2010sun}, Places\cite{zhou2017places}, and Textures\cite{cimpoi2014describing} which are commonly used OOD test datasets. 
These OOD datasets are disjoint from ImageNet-1k classes.

\textbf{Few-shot protocol.}
We sample $K\in\{1,2,4,8,16\}$ labeled images per class from the ID training set to construct the few-shot cache. 
No additional fine-tuning of CLIP is performed. 
For fair comparison, all few-shot methods are optimized using the same class splits and $K$-shot support samples.

\textbf{Implementation Details.}
All experiments are implemented in PyTorch, and the Gaussian Process modules are implemented using GPyTorch. 
All training and evaluation are conducted on a single NVIDIA Tesla V100 GPU with 32GB memory.
All reported results are averaged over three random seeds to ensure robustness.
CLIP ViT-B/16~\cite{dosovitskiy2021vit} is used as the default backbone.
The RBF kernel length-scale $\theta$ is selected via grid search over a range from 0.10 to 2.0 with a step size of 0.05.
Unless otherwise specified, all hyperparameters are fixed to their default values across all experiments.

\textbf{Baselines.}
We compare our method with several existing methods for zero-shot OOD detection, fine-tuned OOD detection, and few-shot OOD detection.
For zero-shot OOD detection, we compare with MCM\cite{ming2022mcm}, GL-MCM\cite{miyai2025glmcm}, and SeTAR\cite{li2024setar}, which are simple yet strong OOD detection methods.
For few-shot OOD detection, we compare with CoOp\cite{zhou2022coop} and LoCoOp\cite{miyai2023locoop}.
For fine-tuned OOD detection, we compare with unimodal methods, MSP\cite{fort2021exploring}, ODIN\cite{liang2018odin}, ViM\cite{wang2022vim}, KNN\cite{sun2022deepnn}, as well as multimodal methods, NPOS\cite{tao2023npos} and CLIPN\cite{wang2023clipn}.

\textbf{Evaluation Metrics.}
For ID classification on the ImageNet-1k dataset, we compute Top-1 accuracy on the ID test set. 
For OOD detection, we treat ID samples as negative and OOD samples as positive. 
We report threshold-free AUROC (higher is better) and FPR95 (false positive rate at 95\% TPR; lower is better). 

\subsection{Experimental Results}
Table~\ref{tab:imagenet-1k-ood} shows OOD detection performance across multiple datasets.
Among few-shot methods, GP-Adapter achieves competitive performance compared to few-shot prompt-learning baselines such as CoOp and LoCoOp, 
despite not relying on gradient-based optimization.

Notably, combining GP-Adapter with prompt-learning methods consistently improves OOD detection performance across datasets.
In particular, GP-Adapter+LoCoOp achieves the best average AUROC and the lowest average FPR95, demonstrating that GP-based uncertainty modeling is complementary to prompt learning.
These results indicate that GP-Adapter can effectively enhance existing few-shot CLIP adaptation methods by modeling predictive uncertainty.
Furthermore, GP-Adapter-based methods outperform zero-shot baselines and are competitive with fine-tuned OOD detection approaches, 
highlighting their effectiveness under data-scarce settings.

In addition to OOD detection, we evaluate ID classification accuracy.
As shown in Table~\ref{tab:id-accuracy}, GP-Adapter preserves competitive ID accuracy compared to existing baselines.
Moreover, integrating GP-Adapter with CoOp improves Top-1 accuracy over the original CoOp method, while its combination with LoCoOp maintains comparable performance.
These results confirm that the proposed method enhances robustness to distribution shifts without sacrificing in-distribution classification accuracy.

\subsection{Ablation Study}
\textbf{Ablation on the Upper Bound $\tau$.}
Table~\ref{tab:ablation-tau} shows the effect of the upper bound $\tau$ that prevents overfitting in few-shot settings, on OOD detection performance.
In this ablation, we use the LoCoOp-learned prompts (GP-Adapter+LoCoOp) and fix the modality weight $\alpha$ to $0.15$, while evaluating $\tau \in \{-10, -5, 0, 5, \infty\}$ on OOD detection performance.
Here, $\tau=\infty$ indicates disabling the upper bound.
$\tau=0$ or $\tau=-5$ yield improved OOD detection performance, achieving lower FPR95 and higher AUROC.

\textbf{Ablation on the ratio of modality $\alpha$.}
Table~\ref{tab:ablation-alpha} shows the effect of the modality weight $\alpha$ that controls the contribution of image and text modalities in the combined predictive distribution. 
In this ablation, we use the LoCoOp-learned prompts (GP-Adapter+LoCoOp) and fix the upper bound $\tau$ to $-5$, while varying $\alpha \in \{0.05, 0.1, 0.15, 0.2\}$.
We observe that $\alpha = 0.15$ yields improved OOD detection performance, achieving lower FPR95 and higher AUROC.

\begin{figure}[t]
  \centering
  \includegraphics[width=0.95\linewidth]{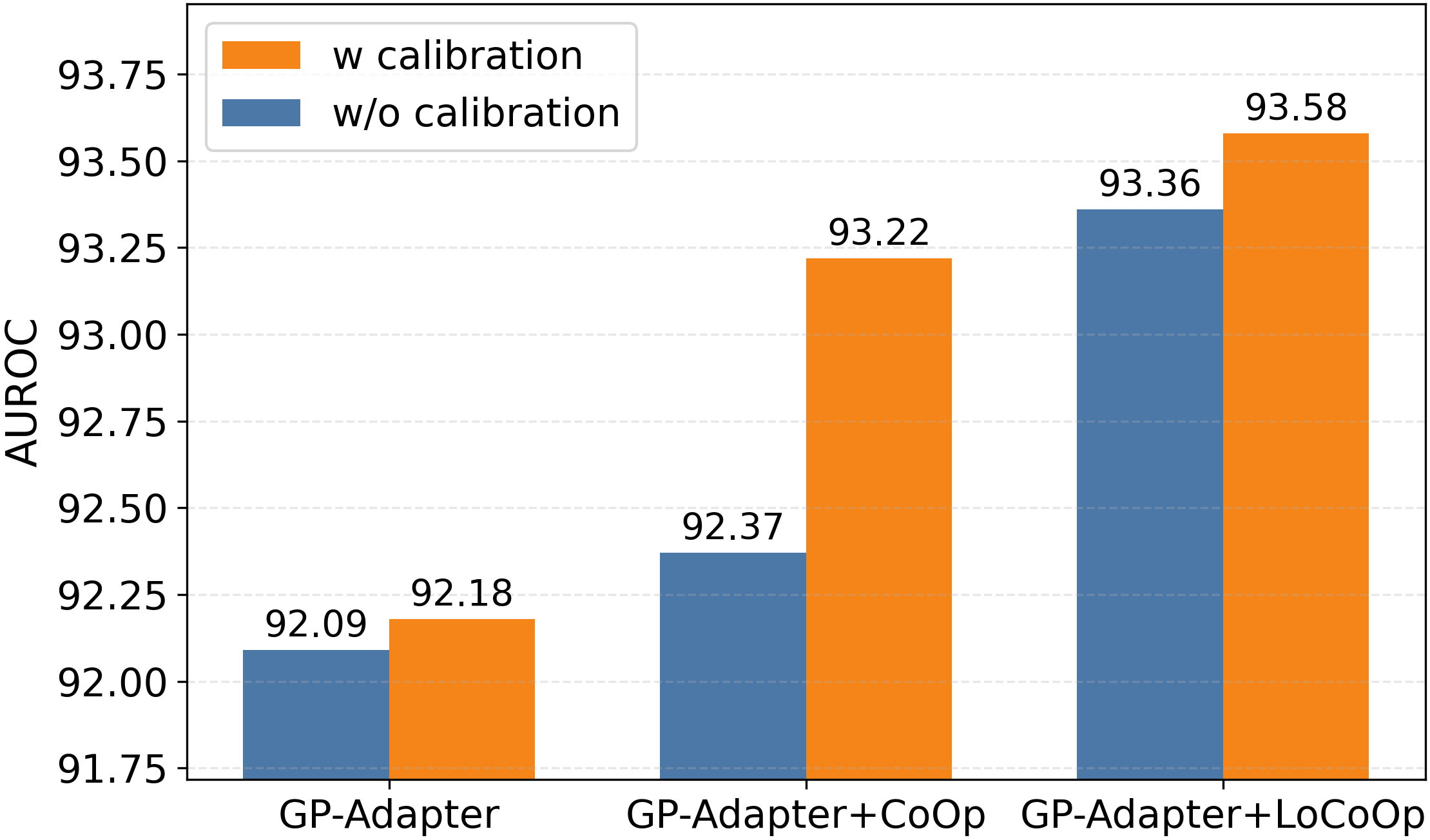}
  \caption{Ablation of variance-aware MSP calibration ($\mathrm{MSP}_{\mathrm{var}}$) for GP-Adapter predictions under different prompt settings.}
  \label{fig:uncertainty-ablation}
\end{figure}

\textbf{Ablation on Variance-Aware MSP Calibration.}
We investigate the effect of incorporating GP predictive variance into the OOD score using the proposed $\text{MSP}_{\text{var}}$ defined in \eqref{eq:mspvar}.
Fig.~\ref{fig:uncertainty-ablation} compares OOD detection performance with and without variance-aware MSP calibration under different prompt settings.
Across all settings, incorporating GP predictive variance consistently improves AUROC, demonstrating the effectiveness of variance-aware MSP calibration.
While the improvement for GP-Adapter alone is modest, substantial performance gains are observed when combined with prompt-learning methods.
In particular, GP-Adapter+CoOp shows a significant increase in AUROC when $\text{MSP}_{\text{var}}$ is applied.
This suggests that variance-aware MSP calibration is especially beneficial when prompt learning sharpens class confidence, 
helping mitigate overconfident predictions.
For GP-Adapter+LoCoOp, performance gains are smaller but still consistent,
indicating that variance-aware calibration provides complementary benefits even when the base method is already strong.

\begin{figure}[t]
  \centering
  \begin{subfigure}{0.48\linewidth}
    \centering
    \includegraphics[width=\linewidth]{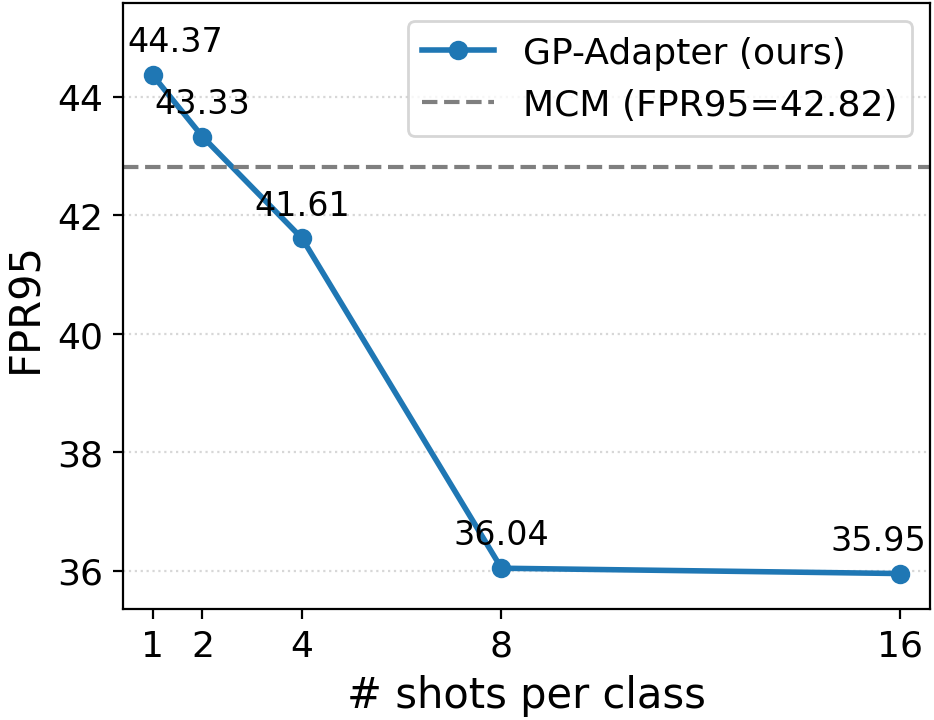}
    \caption{FPR95 vs. shots}
    \label{fig:shots-fpr95}
  \end{subfigure}\hfill
  \begin{subfigure}{0.48\linewidth}
    \centering
    \includegraphics[width=\linewidth]{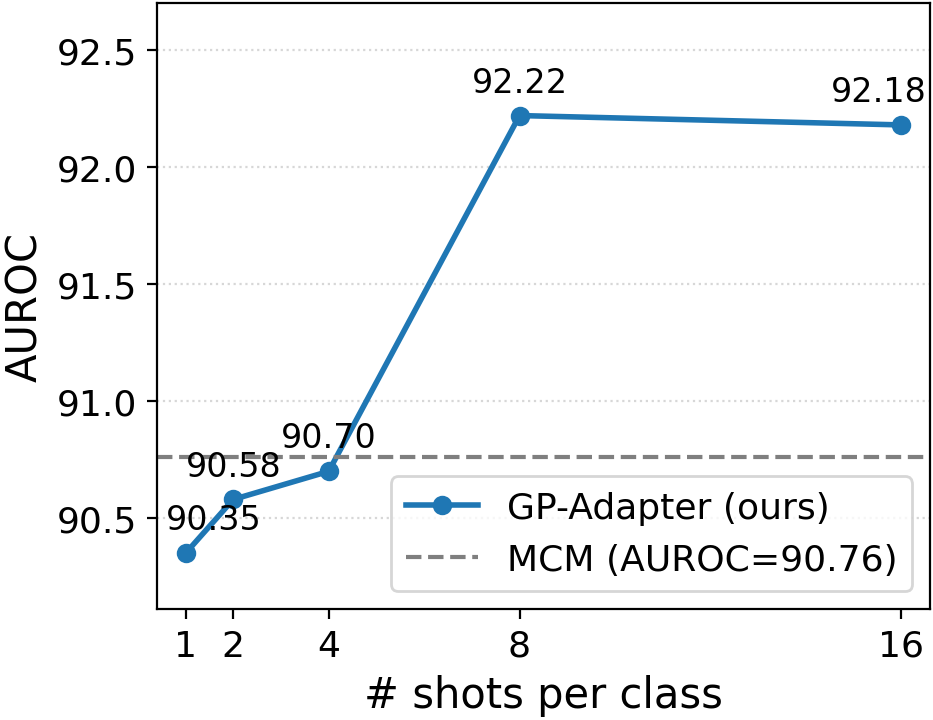}
    \caption{AUROC vs. shots}
    \label{fig:shots-auroc}
  \end{subfigure}
  \caption{Effect of the number of shots per class on OOD detection performance,
compared to MCM baseline.}
  \label{fig:ablation-shots}
\end{figure}

\textbf{Ablation on the Number of Shots.}
Fig.~\ref{fig:ablation-shots} shows the effect of varying the number of shots per class on OOD detection performance.
In this experiment, we fix the modality weight $\alpha$ to $0.15$.
For numerical stability, we set the upper bound $\tau=0$ for the 1-shot and 2-shot settings, and use $\tau=-5$ for all other shot settings.
GP-Adapter exhibits comparable performance to the MCM baseline in extremely low-shot regimes (1,2,4 shots), and begins to consistently outperform it at 8 shots,
with the largest improvement observed between 4 and 8 shots settings, where both FPR95 decreases substantially and AUROC increases.

\textbf{Different Backbone.}
Table~\ref{tab:rn50-ood} shows OOD detection performance using CLIP with a ResNet-50 (RN50) backbone~\cite{he2016resnet}.
GP-Adapter achieves competitive OOD detection performance compared to MCM, CoOp, and LoCoOp.
Moreover, combining GP-Adapter with CoOp or LoCoOp continues to yield performance gains under the RN50 backbone.

\textbf{Small Scale ID Dataset (ImageNet-100).}
We further evaluate GP-Adapter under a reduced ID dataset setting using ImageNet-100, 
following the same set of 100 ImageNet classes as prior work~\cite{tao2023npos}.
Table~\ref{tab:imagenet-100-ood} shows OOD detection performance when the number of ID classes is limited.
Under this setting, GP-Adapter achieves strong OOD detection performance compared to existing baselines.
In particular, GP-Adapter attains the best FPR95 and AUROC on iNaturalist, and maintains competitive average performance across all OOD datasets.
Moreover, combining GP-Adapter with prompt-learning methods continues to provide stable performance.
These results suggest that GP-Adapter remains effective even when the number of ID classes is substantially reduced.

\section{Conclusion}

In this paper, we proposed GP-Adapter, a lightweight adaptation method that augments
frozen CLIP representations with class-wise one-class Gaussian Processes for
few-shot classification and OOD detection.
GP-Adapter avoids gradient-based optimization and relies on closed-form GP
inference with simple kernel hyperparameter selection, allowing rapid
adaptation with memory footprint scaling as $O(CK^2)$ in few-shot settings.
Extensive experiments on ImageNet-1k and multiple OOD benchmarks demonstrate
that GP-Adapter achieves strong and competitive OOD detection performance,
and remains effective when the number of ID classes is substantially reduced or when different backbone architectures are used.
Moreover, GP-Adapter is complementary to prompt-learning methods such as CoOp
and LoCoOp, and their combination yields stable performance gains.
Our ablation analysis further shows that incorporating GP predictive variance
into OOD scoring provides additional improvements.
Future work includes extending GP-Adapter to other few-shot recognition
settings, as well as exploring variational sparse Gaussian Process
approximations to broaden its applicability beyond extremely low-shot regimes.

{\small
\bibliographystyle{IEEEtran}
\bibliography{egbib}
}
\end{document}